\documentclass[conf]{new-aiaa}
\usepackage[utf8]{inputenc}

\usepackage{graphicx}
\usepackage{amsmath}
\usepackage[version=4]{mhchem}
\usepackage{siunitx}
\usepackage{longtable,tabularx}
\usepackage{url}

\newtheorem{definition}{Definition}
\usepackage{algorithm}
\usepackage{algpseudocode}

\title{Wasserstein Distributionally Robust Control Barrier Function using Conditional Value-at-Risk with Differentiable Convex Programming}

\author{Alaa Eddine Chriat\footnote{Ph.D student,  Aerospace Engineering Department, Mississippi State University, Starkville, MS 39759, USA.} and Chuangchuang Sun\footnote{Assistant professor,  Aerospace Engineering Department, Mississippi State University, Starkville, MS 39759, USA.}}

\usepackage{xcolor}

\usepackage{sectsty}

\usepackage{microtype}
\usepackage{graphicx}
\usepackage{subfigure}
\usepackage{booktabs} 
\usepackage{balance}
\usepackage{pdfpages}
\usepackage{makecell}
\usepackage{bm}

\usepackage{flafter}

\usepackage{multirow}
\usepackage{amsmath}
\usepackage{algorithm,algpseudocode}
\usepackage{wrapfig}

\newcommand{\bea}{\begin{eqnarray}}
\newcommand{\eea}{\end{eqnarray}}
\newcommand{\beas}{\begin{eqnarray*}}
\newcommand{\eeas}{\end{eqnarray*}}
\newcommand{\leftm}{\left[\begin{array}}
\newcommand{\rightm}{\end{array}\right]}

\usepackage{diagbox,hhline}

\definecolor{commentclr}{RGB}{110, 149, 204}

\usepackage{stackengine}
\stackMath

\usepackage{color}
\usepackage{soul}
\usepackage{xspace}
\usepackage{cleveref}

\begin{document}

\maketitle

\begin{abstract}
Control Barrier functions (CBFs) have attracted extensive attention for designing safe controllers for their deployment in real-world safety-critical systems. 
However, the perception of the surrounding environment is often subject to stochasticity and further distributional shift from the nominal one. 
In this paper, we present distributional robust CBF (DR-CBF) to achieve resilience under distributional shift while keeping the advantages of CBF, such as computational efficacy and forward invariance.

To achieve this goal, we first propose a single-level convex reformulation to estimate the conditional value at risk (CVaR) of the safety constraints under distributional shift measured by a Wasserstein metric, which is by nature tri-level programming. Moreover, to construct a control barrier condition to enforce the forward invariance of the CVaR, the technique of differentiable convex programming is applied to enable differentiation through the optimization layer of CVaR estimation. We also provide an approximate variant of DR-CBF for higher-order systems. Simulation results are presented to validate the chance-constrained safety guarantee under the distributional shift in both first and second-order systems.
\end{abstract}

\section{Introduction}

\lettrine[]{A}{utonomous} systems are nowadays ubiquitous in the world, from daily life assistance for household chores, and industrial productions, to space explorations, which have significantly changed human society. For example, in 2020, there are 276 million motor vehicles registered in the U.S.\cite{Motor}, and the global market of industrial robots was estimated at around $\$ 55$ billion and is projected to surpass $\$ 165$ billion in 2028 \cite{industrial}. 
However, there can often be perturbations, noise, or malicious attacks during sensing and perception, communication, and actuation. This issue is further exacerbated in unstructured and dynamical environments, such as off-road vehicles, multi-domain operations, and space explorations.

Consider a challenging rescue mission with a robot system after an earthquake. Due to the damage to the cyber-physical infrastructure, communication with the agent can be very limited/ corrupted. Also, the mission can have many unanticipated scenarios, such as road closure. Moreover, in future battlefields, military forces will be deployed in complex environments, including multi-domain operations (MDO) against adversaries by combining the traditional domains (e.g., air, maritime) and the information and electromagnetic domains. In MDO, the goal is to achieve superiority by "connecting distributed sensors, shooters, and data from all domains to joint forces, enabling the coordinated exercise of authority to integrate planning and synchronize convergence in time, space, and purpose" (\cite{USAF}, page 6). In such operations, pervasive uncertainty/perturbation from the environments, intentional/ stealthy/ deceptive adversarial attacks from the opponents, and cyber-physical dysfunctionality can all possibly lead to mission failures. 

In aerial flights and space explorations, such scenarios are also common. A quadcopter, deployed in the open world for wild wire monitoring and disaster relief, can often encounter unexpected gusts. 
Moreover, the CADRE (Cooperative Autonomous Distributed Robotic Exploration) program from NASA JPL aims to achieve collaborative, autonomous exploration, and formation sensing with an integrative pipeline of sensing, perception, communication, computing, and decision-making. Small rovers will be deployed, which will explore the lunar surface, share data, and cooperatively make decisions to eventually accomplish tasks, such as generating a digital elevation map. Moreover, the first Mars helicopter Ingenuity was launched with Mars rovers (e.g., Perseverance) for more effective Mars exploration. Intuitively, the condition is much more restrictive on the Moon/ Mars, regarding atmospheric data, communication quality, the toughness of terrain, and eventually the degradation of all equipment (e.g., perception sensors) on board without maintenance. As a result, the noise and uncertainty from sensing and perception, inter-vehicle communication, planning and control, and actuation will bring significant challenges to collaborative decision-making and coordination. In other words, while a prior of those quantities of interest is available, such probabilistic distribution can shift. 
In summary, decision-making in autonomous systems with rigorous robustness and resilience is highly desired in many applications, especially those with humans in the loop.
To this end, our overarching goal is to develop a distributed approach for distributionally robust control and decision-making for autonomous systems with applications in space exploration.


Due to the importance of decision-making under distributional shift, there exist many works of distributionally robust control for a single agent in the framework of model predictive control (MPC), mostly also using chance constraints such as conditional value-at-risk \cite{van2015distributionally, hakobyan2021wasserstein, bahari2022safe, coulson2021distributionally, coppens2021data}. There are also works based on approximate dynamic programming to achieve distributional robustness \cite{yang2020wasserstein}. 
However, those methods often require solving a complex optimization problem, possibly in the minimax form, making it inapplicable for online efficient control. 
Comparatively, control barrier functions(CBF~\cite{ames2016control}) have attracted much attention with diverse variants in different settings. One of the reasons is their computational efficacy, which only requires successively solving a convex quadratic programming regarding one time step for general control affine systems. Moreover, the forward invariance of safety satisfaction is also guaranteed via the control barrier condition. However, exactly combining chanced-constrained distributional optimization and control barrier function is not straightforward and we aim to bridge the gaps. 

Consider a safety constraint $h(x,w)\le0$, where $x$ is the state and $w$ is the noise subject to distributional shift. The chance-constrained safety specification, $\mathrm{CVaR} \circ h(x,w)$, is often estimated by conditional value at risk (CVaR) by solving optimization problems. Here ``$\circ$" denotes function composition. It comes to the first issue: estimating the CVaR under distributional shift is nontrivial; it is by nature a tri-level problem: CVaR optimization and the primal-dual optimization considering the distributional constraint. To address this issue, we present an approximate approach to keep tractability while avoiding over-conservatism, which eventually leads to single-level convex programming.
Subsequently, it comes to the forward invariance and satisfaction of the $\mathrm{CVaR} \circ h(x,w)$ by control barrier functions.
Naturally, the control barrier condition (CBC) that enforces the forward invariance of the safety constraints under distributional shift is in the form of $\mathrm{CBC} \circ \mathrm{CVaR} \circ h(x,w)$. It is known that CBC needs the differentiation of its argument, which in this case $\mathrm{CVaR} \circ h(x,w)$. However, with $\mathrm{CVaR}(\bullet)$ as an optimization layer, it is not immediately clear how to differentiate through it to construct the CBC. To circumvent this issue, work in \cite{long2022safe} estimates the conditional value at risk of the control barrier condition, instead of the CVaR estimate of the original constraint $h(x,w)$. That is to say, a relaxed criterion is imposed by enforcing the chance-constrained control barrier condition (i.e., $\left.\mathrm{CVaR}_\alpha \circ \mathrm{CBC} \circ h(x,w)\right)$, instead of enforcing the forward variance of the real chance-constrained safety constraint (i.e., $\mathrm{CBC} \circ {\mathrm{CVaR}}_\alpha \circ h(x,w)$ ). As a result, $\mathrm{CVaR}_\alpha \circ \mathrm{CBC} \circ$ $h(x,w)$ only needs to differentiate $h(x,w)$ itself, which is much easier compared to the differentiation through the optimization layer $\mathrm{CVaR}_\alpha \circ h(x,w)$ over $x$.

\subsection{Contributions}


To enable distributional robustness while keeping the advantages of the control barrier function, we make the following contributions to bridge the aforementioned gaps with distributionally robust control barrier functions.
\begin{itemize}
    \item  We present a formulation to simply the tri-level optimization problem for estimating the $\mathrm{CVaR}$ into a single-level convex program, then use differentiable convex programming to enable differentiation through the optimization layer $\mathrm{CVaR} \circ h(x,w)$
    \item We construct the control barrier function to enforce the forward invariance of the chance-constrained safety constraint (i.e., $\mathrm{CVaR} \circ h(x,w)$). As a result, it makes it possible to enforce $\mathrm{CBC} \circ \mathrm{CVaR}_\alpha \circ h(x,w)$ that can capture the essence of the problem and eventually guarantee the forward invariance and the satisfaction of the safety specification. We also provide an approximate method for higher-order systems.
    \item We present simulation results where the strength of the proposed distributionally robust control barrier function (i.e., remaining safe under distributional shift) in stochastic environments is demonstrated compared to the vanilla CBFs. 
\end{itemize}

\subsection{Related Works}

There have been extensive existing works\cite{madry2017towards,lutjens2020certified} studying robust learning and control under perturbations, often formulated as $l_p$ balls as $\left\|x-x_0\right\|_p \leq \epsilon$, with $x$ and $x_0$ as the quantity of interest and its nominal value respectively and $\epsilon$ as the radius of the $l_p$ ball. More specifically, the perturbations in observations \cite{liu2022robustness}, action \cite{tessler2019action}, model \cite{mankowitz2019robust}, and those in the context of safe reinforcement learning (RL)\cite{brunke2022safe,achiam2017constrained} have been well studied. In multi-agent RL, model uncertainty\cite{zhang2020robust1}, adversary agents \cite{zhang2020robust1, sun2022romax}, and beyond have been considered to achieve robustness. 
Correspondingly, the distributional shift defined as $x\sim p(x), d(p(x), p_0(x))\le\epsilon$, where $p(x)$ and $ p_0(x)$ are the real and nominal probability distributions of $x$ with $d(\bullet,\bullet)$ as a distance measurement between two probability distributions and $\epsilon$ the threshold. Compared to the $l_p$ ball perturbation, the distributional shift admits a much larger space to explore to find the worst adversarial behaviors, which the agent is expected to mitigate in a principled way.
For distributionally robust single-agent learning and control under distributional shift, model-based approaches\cite{van2015distributionally} such as approximate dynamic programming \cite{yang2020wasserstein,hakobyan2021wasserstein} and model predictive control \cite{bahari2022safe, coulson2021distributionally, coppens2021data} have been proposed, with a chance-constrained criterion under a Wasserstein metric. Specifically, in the framework of model predictive control, many existing works of distributionally robust control use conditional value-at-risk \cite{van2015distributionally, hakobyan2021wasserstein, bahari2022safe, coulson2021distributionally, coppens2021data}. In the model-free regime, one line of work is to generate environments/ tasks with distributional shifts, for policy training to achieve robustness\cite{ren2022distributionally, morrison2020egad, wang2019adversarial}. Moreover, to balance the worst-case (robustness) and average performance, \cite{xu2022group} trains policies over task groups by adding regularization to the worst possible outcomes.  Additionally, with the presence of adversaries, the minimax/ bi-level optimization formulation~\cite{yang2022stackelberg} is often employed for worst-case robustness, which is challenging to solve.

\subsection{Organization}
To ensure that the research provides a comprehensive understanding of the topic at hand, this research paper is organized into four main sections. The first section introduces the background and significance of the research, highlighting the motivation behind the study and its relevance. The rest of this paper is organized as follows. In Section II, we revisit some fundamental concepts and definitions of control barrier functions, differentiable convex programming, and distributionally robust optimization. 
In section III, we start by developing the optimization problem to estimate $\mathrm{CVaR}$ under distributional shift and calculate the relevant gradients. Then we construct the control barrier function of the $\mathrm{CVaR}$ for first-order systems and provide an approximate method for higher-order systems.
In section IV, we present simulation results for first-order and second-order systems highlighting the advantage of using the Distributionally Robust Control Barrier Function(DR-CBF).
Finally, section V summarizes our research, and discusses some future ideas that warrant exploration. 

\section{Preliminary}

\subsection{Control Barrier Function}

Control Barrier Functions (CBFs) are mathematical tools used to ensure the safety and stability of dynamic systems. A CBF is a function that maps the current state of the system to a value that measures how far the system is from violating the desired safety constraint. The control law is then designed to enforce the CBF such that it remains within the safe region.
The control law is typically designed using a Lyapunov-based approach, where a Lyapunov function is chosen to drive the system to the desired behavior and solved in the form of a Quadratic Program (QP), while The barrier function is incorporated into the QP to guarantee the desired safety specifications.
Mathematically, consider the nonlinear control-affine system:

\begin{equation}\label{nonlinearsystem}
\dot{x}(t)=f(x(t))+g(x(t)) u(t)
\end{equation}
where $f$ and $g$ are globally Lipschitz, $x\in\mathbb{R}^n$ and $u\in\mathbb{R}^m$ are the states and control inputs, respectively, constrained in closed sets, with initial condition $x(t_0) = x_0$.

\begin{definition}
\cite{ames2016control}$h: \mathbb{R}^{n} \rightarrow \mathbb{R}$ is a barrier function for the set $C=\left\{x \in \mathbb{R}^{n}: h(x) \geqslant 0\right\}$ if $\exists$ $\mathcal{K}$ function $\alpha(\bullet)$ such that:
\begin{equation}
\begin{gathered}
\sup _{u \in U}[L_{f}h(x)+L_{g}h(x) u+\alpha(h(x))] \geqslant 0 \\
\inf_{\text{int}(C)}[\alpha(h(x)) ]\geqslant 0 \text {  \quad  and  \quad  } \lim_{\partial C} \alpha(h(x))=0
\end{gathered}
\end{equation}
\end{definition}
Because not all systems are first-order in inputs, we can use higher-order control barrier functions to constrain higher-order systems.
\begin{definition}\label{def:HOCBF}
\cite{xiaohigh}For the non linear system \eqref{nonlinearsystem} with the $m^{th}$ differentiable function $h(x)$ as a constraint, we define a sequence of functions $\psi_{i}$ with $i \in \{1,2,...,m\}$, starting from $\psi_{0}=h(x)$:
\begin{equation}\label{Hcbfform}
\psi_{i}(x, t)=\dot{\psi}_{i-1}(x, t)+\alpha_{i}\left(\psi_{i-1}(x, t)\right) 
\end{equation}
the function $h(x)$ is a high order control barrier function if $\exists$ $\mathcal{K}$ functions $\alpha_{i}(\bullet)$ such that:
\begin{equation}
\psi_{m}(x, t) \geqslant 0 
\end{equation}
\end{definition}

By solving the following Quadratic Program, leveraging the power of CBF, we can enforce safety constraints, maintain stability, and prevent undesirable behavior  
\begin{equation}\label{eq:minnorm}
\begin{aligned}
\min _{u\in[\underline{u}, \bar{u}]} \ \ \ & J(x,u) \\
\text { s.t. } \ \ \ & \frac{\partial h(x)}{\partial x}(f(x)+g(x) u)+\kappa(h(x))\geq 0 \\
\end{aligned}
\end{equation}

\subsection{Differentiable Convex Programming}\label{sec:DP}

Differentiable convex programming is a powerful technique that enables the computation of gradients for the objective function of an optimization problem with respect to its parameters. This is achieved by applying matrix differentiation to the Karush-Kuhn-Tucker (KKT) conditions. A notable example of a differentiable optimization method is OPTNET ~\cite{amos2017optnet}, which incorporates differentiable optimization problems within the architecture of neural networks. During training, the gradients of the objective function are computed and back-propagated through the network. In a broader sense, this methodology can be applied to differentiate through disciplined convex programs ~\cite{diffcvxoptlayer} by initially mapping them into cone programs ~\cite{diffconeprog}, computing the gradients, and subsequently mapping back to the original problem. A common application of differentiable programming is learning the constraints of an optimization problem, such as convex polytopes or ellipsoid projections. The major advantage of differentiable optimization methods, such as OPTNET, lies in their ability to optimize a wide range of challenging convex objectives that are typically difficult to handle using traditional optimization approaches. It is worth noting that the convex quadratic program (QP) in equation \eqref{eq:minnorm} can be differentiated through the KKT conditions \cite{amos2017optnet}, which serve as equivalent conditions for global optimality. According to the KKT conditions, at the optimal solution, the gradient of the Lagrangian function with respect to the program's input and parameters must be zero. Consequently, by taking the partial derivative of the Lagrangian function with respect to the input and extending it through the chain rule to the program's parameters, their gradients can be obtained. We have integrated differentiable optimization using the cvxpylayers package \footnote{\url{https://github.com/cvxgrp/cvxpylayers}} which is an extension to the cvxpy package with an affine-solver-affine (ASA) approach. The ASA consists of taking the optimization problem's objective and constraints and mapping them to a cone program.
For a generalized QP:
\begin{equation}
\begin{aligned}
\min_x &\ \  \frac{1}{2} x^T Q x+q^T x \\
\text { s.t.} &\ \  A x=b \\
&\ \   G x \leq h,
\end{aligned}
\end{equation}
we can write the Lagrangian of the problem as:
\begin{equation}
L(z, \nu, \lambda)=\frac{1}{2} z^T Q z+q^T z+\nu^T(A z-b)+\lambda^T(G z-h)
\end{equation}
where $\nu$ are the dual variables on the equality constraints and $\lambda \geq 0$ are the dual variables on the inequality constraint.
Using the KKT conditions for stationarity, primal feasibility, and complementary slackness.
\begin{equation}
\begin{aligned}
Q z^{\star}+q+A^T \nu^{\star}+G^T \lambda^{\star} & =0 \\
A z^{\star}-b & =0 \\
D\left(\lambda^{\star}\right)\left(G z^{\star}-h\right) & =0
\end{aligned}
\end{equation}
By differentiating these conditions, we can shape the Jacobian of the problem as follows.
\begin{equation}
\left[\begin{array}{l}
d_z \\
d_\lambda \\
d_\nu
\end{array}\right]=-\left[\begin{array}{ccc}
Q & G^T D\left(\lambda^{\star}\right) & A^T \\
G & D\left(G z^{\star}-h\right) & 0 \\
A & 0 & 0
\end{array}\right]^{-1}\left[\begin{array}{c}
\left(\frac{\partial \ell}{\partial z^{\star}}\right)^T \\
0 \\
0
\end{array}\right]
\end{equation}
Furthermore, via chain rule, we can get the derivatives of any loss function of interest regarding any of the parameters in the QP.

\subsection{Distributionally Robust Optimization and Conditional Value at Risk}

Distributionally Robust Optimization (DRO) is an approach to optimization under uncertainty that aims to find solutions that are robust against a wide range of possible probability distributions. Unlike traditional optimization methods that assume a known probability distribution for uncertain parameters, DRO takes a more cautious approach by considering a set of possible distributions and optimizing for the worst-case scenario within that set.
In DRO, the uncertain parameters are typically modeled as random variables with unknown distributions. The goal is to find a solution that performs well across all possible distributions within a certain ambiguity set. The ambiguity set represents the range of possible distributions and is defined based on certain metrics, such as statistical moments or the Wasserstein metric.
The key idea in DRO is to find a solution that minimizes the expected cost or maximizes the expected reward under the worst-case distribution within the ambiguity set. This approach provides a robust solution that performs well regardless of the actual distribution of the uncertain parameters.
DRO has applications in various domains, including operations research, finance, and machine learning. It can be used to optimize decisions in settings with uncertain data, such as supply chain management, portfolio optimization, or predictive modeling with uncertain inputs.
However, solving DRO problems can be challenging due to the increased complexity introduced by worst-case optimization. The optimization problem typically becomes non-convex and computationally demanding. Various techniques, such as convex relaxations, scenario approximations, or sample-based methods, are used to handle the computational challenges associated with DRO.
In general, the value at risk of a random quantity $h(x,w)$ (with $h$ as the shorthand notation) with a confidence level of $\alpha$ is : 
\begin{equation}\label{eq:var}
\operatorname{VaR}_\alpha(h):=\min \{\eta \in \mathbb{R} \mid \mathbb{P}(h \leq \eta) \geq \alpha\}
\end{equation}
which can be interpreted as the worst-case scenario risk with probability $\alpha$. Due to the complexity of solving for the VAR, we define a more efficient version, the conditional value at risk which can be formulated as the following convex program:
\begin{equation}\label{CVaR}
\mathrm{CVaR}_\alpha(h):=\min _{\eta \in \mathbb{R}} \mathbb{E}\left[\eta+\frac{(h-\eta)_{+}}{1-\alpha}\right]
\end{equation}
which can be subsequently reformulated into a tractable linear program: 
\begin{equation}
\begin{aligned}
\mathrm{CVaR}_\alpha \left(h\right) \approx
& \min _{\eta_i, s_i} \quad \eta_i+\frac{1}{(1-\alpha) N_s} \sum_{m=1}^{N_s} s_i^m \\
& \text { s.t. }\quad h-\eta_i \leq s_i^m, \quad \forall m \in \mathbb{I}_{1: N_s} \\
& \quad\quad\quad0 \leq s_i^m, \quad \forall m \in \mathbb{I}_{1: N_s}
\end{aligned}
\end{equation}
where $I \in \mathbb{N}$ is the set of estimated samples. In order to take into account the unmeasured distributions, we introduce the Wasserstein metric and build an ambiguity set. This enables solving the problem for the worst-case scenario. For all distributions $\mathcal{P}_1, \mathcal{P}_2 \in \mathcal{P}(\mathbb{W})$ we can define the Wasserstein metric as: 
\begin{equation}
d_{\mathrm{W}}\left(\mathcal{P}_1, \mathcal{P}_2\right):= \min _{\kappa \in \mathcal{P}\left(\mathbb{W}^2\right)}\left\{\int_{\mathbb{W}^2}\left\|w_1-w_2\right\| \mathrm{d} \kappa\left(w_1, w_2\right)\right. 
\left.\times \mid \Pi^l \kappa=\mathcal{P}_l, l=1,2\right\}
\end{equation}
integrating the Wasserstein metric into the $\mathrm{CVaR}$ linear program to optimize over the whole ambiguity set results in the following optimization problem \cite{bahari2022safe}:
\begin{equation}
\sup _{\mathcal{P} \in \mathbb{D}} \mathrm{CVaR}_\alpha^{\mathcal{P}}(h)=\inf _{\lambda \geq 0}\left\{\lambda \epsilon+\frac{1}{N_s} \sum_{m=1}^{N_s} \sup _{w \in \mathbb{W}}\left\{[h-\eta]_{+}-\lambda\left\|w-w^m\right\|\right\}\right\}
\end{equation}
Overall, distributionally robust optimization provides a principled approach to decision-making under uncertainty, offers robustness guarantees, and can lead to more reliable and resilient solutions in uncertain environments.

\section{Chance Constrained Distributionally Robust Control Barrier Functions with a Wasserstein Metric}
\subsection{The Estimation of Conditional Value-at-Risk under Distributional Shifts: A Simplified Formulation} 
Consider a general dynamical system in the following form
\begin{equation}\label{eq:dynamics}
    \dot{x}=f\left(x, u\right)
\end{equation}
where $x \in \mathbb{R}^{n}, u \in \mathbb{R}^{m}, f: \mathbb{R}^{n+m} \rightarrow \mathbb{R}^{n}$ are the state, control input, and the dynamical transition function, respectively. Naturally, in many scenarios, there arise safety constraints, such as obstacle avoidance formulated as $h(x) \leq 0$. Without loss of generality, only one constraint is considered here and our approach can be easily extended to multiple-constraint cases.

As the dynamics in \eqref{eq:dynamics} are deterministic, we consider stochastic constraints with additive noise $h(x,w) \leq 0$. As a result, it is desirable to satisfy the constraint with as high a probability as possible, with a prerequisite first to estimate the worst case $h(x,w)$ under stochasticity. Then value-at-risk and its more tractable approximation conditional value-at-risk (CVaR) \cite{rockafellar2000optimization} are often used to measure the risks. Mathematically, $\lim _{\alpha \rightarrow 1} \mathrm{CVaR}_\alpha(h(x,w)) \leq$ 0 means that the constraint is satisfied with a probability of at least $\alpha$ (i.e., $\mathbb{P}(h(x,w) \leq 0) \geq \alpha$ ). With $N_s$ independent and identically distributed (i.i.d.) samples of the disturbance $\left\{w^m\right\}_{m=1}^{N_s}$, we can get the corresponding samples of $x$ at the current time step based on the dynamics \eqref{eq:dynamics} and the control input $u$ of last time-step in an online setting. Time dependency is omitted, as it is applicable for all time instances. Then the $\mathrm{CVaR}_\alpha$ can be estimated by solving the following linear programming \cite{rockafellar2000optimization} with auxiliary variables $s^m$
\begin{equation}\label{eq:lincvar}
    \min _{\eta, s}\left\{\eta+\frac{1}{(1-\alpha) N_s} \sum_{m \in\left[N_s\right]} s^m \mid \text { s.t. } h\left(x^m,w^m\right)-\eta \leq s^m, s^m \geq 0, \forall m \in\left[N_s\right]\right\},
\end{equation}
where $\left[N_s\right]=1, \ldots, N_s$ is the index set for the samples.

To estimate the risk constraint $\mathrm{CVaR}_\alpha(h(x,w))$ via \eqref{eq:lincvar}, there often requires many samples (i.e., large $N_s$ ) of the random variables, such as sample average approximation (SAA \cite{kleywegt2002sample}) in data-driven stochastic optimization. However, collecting samples of the disturbance for the physical robotic systems is too restrictive and unsafe, especially with humans in the loop. As such, limited samples will usually not be able to capture the proper distribution of the stochastic variable, admitting an ambiguity set. In other words, the actual distribution might shift from the estimated one. To effectively ensure robustness under such a distributional shift, Distributionally Robust Optimization (DRO \cite{sinha2017certifiable,blanchet2019quantifying,rahimian2019distributionally} will be employed to solve the stochastic optimization problem by considering the worst case within the ambiguity set. Intuitively, the metrics to measure the distance between two probabilistic distributions are used to parameterize the ambiguity set, including the Kullback-Leibler divergence \cite{hu2013kullback} and the Wasserstein metric \cite{givens1984class}. Here the latter is adopted, as distributional robust optimization provides a probabilistic guarantee of out-of-sample performance under a Wasserstein metric \cite{mohajerin2018data}. In the following, we will show how $\mathrm{CVaR}_\alpha$ can be estimated with the Wasserstein ambiguity set.

Denote $p_0(w)$ as the empirical distribution of the random variables $w$ estimated from samples $\left\{w^m\right\}_{m=1}^{N_s}$. Then the ambiguity set of the perturbed distribution $p(w)$ from the nominal distribution $p_0(w)$ under a Wasserstein metric \cite{givens1984class} is expressed as $\mathcal{P}=\left\{p(w) \mid W_d\left(p(w), p_0(w)\right) \leq \rho\right\}$, with $W_d(\bullet, \bullet)$ as the Wasserstein distance and $\rho$ as the threshold of such a shift. To satisfy the constraint, we have the following worst-case scenario in the ambiguity set as $H(x):=\sup _{p \in \mathcal{P}} \operatorname{CVaR}_\alpha^p(h(x,w)) \leq 0$. Then based on the definition of $\mathrm{CVaR}_\alpha$, it can be further reformulated as

\begin{equation}
\begin{aligned}\label{eq"qpcvar}
\sup _{p \in \mathcal{P}} \operatorname{CVaR}_\alpha^p(h(x,w)) \leq & \min _\eta\left\{\eta+\frac{1}{1-\alpha} \sup _p\left\{[h(x,w)-\eta]_{+}-\lambda W_d\left(p(w), p_0(w)\right)\right\}\right\} \\
= & \min _\eta\left\{\eta+\frac{1}{1-\alpha} \frac{1}{N_s} \sum_{m \in\left[N_s\right]} \sup _{w}\left\{[h(x,w)-\eta]_{+}-\lambda\left\|w-w^m\right\|\right\}\right\}
\end{aligned}
\end{equation}
where $[\bullet]_{+}=\max (0, \bullet)$. In the first inequality, we take the ambiguity set constraint as a penalty with $\lambda>0$ to reduce one layer of the optimization problem (i.e., eliminating the minimization over $\lambda$). However, the inner maximization is on the infinite-dimensional probability measure of $w$ and is hence intractable. With Kantorovich duality \cite{mohajerin2018data}, it is further equivalently reformulated as an optimization problem on the finite space of $w$ in the equality. Note that $[\bullet]_{+}$can be transformed as linear constraints with slack variables as \eqref{eq:lincvar}, resulting in an equivalent convex quadratic programming (QP) problem from \eqref{eq"qpcvar}. 
Note that the problem in \eqref{eq"qpcvar} will lead to a bi-level optimization problem if combined with optimal control design. Due to the difficulty of solving bi-level optimization, most existing works plainly list the constraints in \eqref{eq"qpcvar} by removing the minimization over $\eta$ in the overall optimal control problem. This lead to conservatism (e.g., replacing $min_x f(x)\le0$ by $f(x)\le0$). Here we will address this issue principally to keep tractability while mitigating over-conservatism, by efficiently solving the supremum problem in \eqref{eq"qpcvar} to further reduce it from a bi-level problem to a single-level convex programming. To achieve this, it is further assumed that the noise is additive as $h(x,w) = h(x) + w$, where $w\in\mathbb{R}^1$ is in a closed convex set subject to a Gaussian distribution. This can lead to the analytical solution for the supremum problem. with two cases considering the $[\bullet]_+$ operator.

\noindent\textbf{Case 1: $h(x) + w - \eta \leq 0$.}\\
\noindent In this case, the optimal solution for the $\sup _{w}\left\{-\lambda\left\|w-w^m\right\|\right\}$ regarding $w$ is achieved with $w^* = w^m$. Hence, it results in the following linear program:
\begin{equation}\label{eq:1stcase}
\begin{aligned}
\sup _{p \in \mathcal{P}} \operatorname{CVaR}_\alpha^p(h(x,w)) = \min _{\eta} & \left\{\eta+\frac{1}{1-\alpha} \frac{1}{N_s} \sum_{m \in\left[N_s\right]}s^m\right\} \\
\text { s.t. } &  h(x) + w^* - \eta \leq s^m,\\
&h(x) + w^* - \eta \leq 0, \\
&0 \leq s^m. 
\end{aligned}
\end{equation}
In this case, $s^m$ will be zeros (due to the minimization and nature of the first two constraints). Then the estimation of $\sup _{p \in \mathcal{P}} \operatorname{CVaR}_\alpha^p(h(x,w))$ will be the worst case of $h(x) + w^m$, intuitively leading to conservatism under the distributional shift of $w$.

\noindent\textbf{Case 2: $h(x) + w - \eta \geq 0$.}\\
\noindent In this case, the optimal solution for the linear program $ \sup _{w} h(x,w)-\eta-\lambda\left\|w-w^m\right\|$ regarding $w$ is achieved at the vertices of the polytope feasible sets formulated via the linear constraints, including the bounds. Therefore, the value $w^m$, or the bounds of the set $\underline{w},\Bar{w}$ are possible solutions for the supremum operator. As a result, we can arrive at the following linear program: 
\begin{equation}\label{eq:2ndcase}
\begin{aligned}
\sup _{p \in \mathcal{P}} \operatorname{CVaR}_\alpha^p(h(x,w)) = \min _{\eta} & \left\{\eta+\frac{1}{1-\alpha} \frac{1}{N_s} \sum_{m \in\left[N_s\right]}(s^m + L^m)\right\} \\
\text { s.t. }\\
&\left\{\begin{array}{l} h(x)+ \Bar{w} - \eta \leq s^m,\\
h(x)+ \Bar{w} - \eta - \lambda(\Bar{w} - w^m) \leq L^m, \\
h(x)+ \Bar{w} - \eta \geq 0,\\
\end{array}\right. \\
&\left\{\begin{array}{l} h(x)+ \underline{w} - \eta \leq s^m,\\
h(x)+ \underline{w} - \eta + \lambda(\underline{w} - w^m) \leq L^m,\\
h(x)+ \underline{w} - \eta \geq 0,\\
\end{array}\right. \\
&\left\{\begin{array}{l} h(x)+ w^m - \eta \leq s^m,\\
h(x)+ w^m - \eta \leq L^m, \\
h(x)+ w^m - \eta \geq 0,
\end{array}\right. \\
\end{aligned}
\end{equation}
For each potential solution, we get a set of constraints to satisfy in \eqref{eq:2ndcase}. For the term with absolute value operator, i.e., $-\left\|w-w^m\right\|$, it is further reduced as  $(\underline{w} - w^m)$ and $- (\Bar{w} - w^m)$, with $\underline{w}$ and $\Bar{w}$ as the lower and upper bound of $w$, respectively. Also, since $h(x) + \underline{w} - \eta \geq 0$ and $h(x)+ \underline{w} - \eta \leq s^m$, the constraint $0 \leq s^m$ becomes trivial and thus can be removed. Note that the problem in \eqref{eq"qpcvar} is only to estimate the risk constraint and it will lead to a bi-level optimization problem if combined with optimal control design. Due to the difficulty of solving bi-level optimization, most existing works just plainly list the objective (without the minimization) and constraints in \eqref{eq"qpcvar} as extra constraints over $\eta$ in the overall optimal control problem. This lead to conservatism (e.g., replacing $\min_\eta f(\eta)\le0$ by $f(\eta)\le0$). Here we will address this issue principally to keep tractability while mitigating over-conservatism by integrating it with control barrier functions. As a result, we will present how we can use the \textit{optimal value} of $\sup _{p \in \mathcal{P}} \operatorname{CVaR}_\alpha^p(h(x,w))$ and its derivative to construct the control barrier conditions in the following section.


\subsection{Distributionally Robust Control Barrier Functions via Differentiable Convex Programming}
Consider the following non-linear control-affine system $\dot{x}=f(x)+g(x) u$, where $f$ and $g$ are locally Lipschitz, $x \in D \subset \mathbb{R}^n$ is the state and $u \in U \subset \mathbb{R}^m$ is the set of admissible inputs. The safety set is defined as $\mathcal{C}=\left\{x \in D \subset \mathbb{R}^n \mid h(x,w) \leq 0\right\}$ with $\mathcal{C} \subset D$. Then $h$ is a zeroing control barrier function (CBF) \cite{ames2016control} if there exists an extended class- $\kappa_{\infty}$ function $\kappa$ such that for the above control system
\begin{equation}\label{eq:cbf}
    \sup _{u \in U}\left(L_f h(x,w)+L_g h(x,w) u+\kappa(h(x,w))\right) \leq 0, \forall x \in D
\end{equation}
where $L_f h(x)=\left(\frac{\partial h(x,w)}{\partial x}\right)^T f(x)$ is the Lie derivative. Note that $h(x,w) \leq 0$, instead of $h(x,w) \geq 0$ in the CBF literature, defines the safety set for consistency here. As such, it is " $\leq$ " rather than " $\geq$ " in \eqref{eq:cbf}. The control barrier condition (CBC) in \eqref{eq:cbf} will ensure the forward invariance of the constraint $h(x,w)$ and has been extensively studied with many variants. Forward invariance means that the violation of the safety constraint will only become smaller and smaller if starting outside the safety set, and will remain inside otherwise. 

In terms of distributionally robust CBF, work in \cite{long2022safe} estimates the conditional value at risk of the control barrier condition in \eqref{eq:cbf}, instead of the CVaR estimate of the original constraint $h(x,w)$ as in \eqref{eq"qpcvar}. That is to say, the former is applying a relaxed criterion by enforcing the chance-constrained control barrier condition (i.e., $\left.\mathrm{CVaR}_\alpha \circ \mathrm{CBC} \circ h(x,w)\right)$, instead of enforcing the forward variance of the real chance-constrained safety constraint (i.e., $\mathrm{CBC} \circ {\mathrm{CVaR}}_\alpha \circ h(x,w)$ ). Here we use " $\circ$ " to denote function composition to avoid many layers of parentheses. However, while $\mathrm{CBC} \circ \mathrm{CVaR}_\alpha \circ h(x,w)$ can capture the essence of the problem, it brings new challenges. As to differentiate through the optimization layer $\mathrm{CVaR}_\alpha \circ h(x,w)$ over $x$ (non-trivial), while $\mathrm{CVaR}_\alpha \circ \mathrm{CBC} \circ$ $h(x,w)$ only needs to differentiate $h(x,w)$ itself (much easier). we combine distributionally robust control with control barrier functions to enforce the forward invariance of the risk estimate $\mathrm{CVaR}_\alpha \circ h(x,w)$. 
As discussed before, the estimate of $\mathrm{CVaR}_\alpha(h(x,w))$ in \eqref{eq"qpcvar} is a convex quadratic program, which we need to differentiate through over $x$ to construct the control barrier condition in \eqref{eq:cbf}. 
Leveraging recent advances in differentiable convex optimization  \cite{amos2018differentiable,agrawal2019differentiating,agrawal2020learning}, we can formulate our problem as a disciplined parameterized program and use the \textit{cvxpylayers} package to map our problem into a cone program and differentiate the KKT conditions at the optimal solution to get the partial derivatives of the solution of the $\mathrm{CVaR}$ with respect to the problem's parameters; see details in \Cref{sec:DP}. As a result, we are able to calculate $\frac{\partial \mathrm{CVaR}_\alpha(h(x,w))}{\partial x:}$ and construct the control barrier condition as follows
\begin{equation}\label{eq:cvarcbf}
\min _{u}\left\{J\left(x, u\right) \mid \text { s.t. } \frac{\partial \mathrm{CVaR}_\alpha(h(x,w))}{\partial x}\left(f\left(x\right)+g\left(x\right) u\right)+\kappa\left(\operatorname{CVaR}_\alpha(h(x,w))\right) \leq 0, \forall m \in\left[N_s\right]\right\},
\end{equation}
where $J\left(x, u\right)$ is the loss function, such as reference trajectory tracking (e.g., $\left\|x-x_{ref}\right\|_2^2$ ), a Lyapunov function for goal-reaching, optimal fuel consumption (e.g., $\left\|u\right\|_2^2$ ), etc. Problem \eqref{eq:cvarcbf} is often a convex quadratic programming, with a quadratic loss function $J\left(x, u\right)$ as exemplified and a linear CBC constraint inheriting from the general CBF context in \eqref{eq:cbf}. Note that the dynamics are not explicitly included in the optimization, and thus not variables, as CBF only considers a single time-step forward. The single-step system propagation will follow the dynamics in \eqref{eq:dynamics} to get the next state with the control input $u$ from \eqref{eq:cvarcbf}.
In this way, we implicitly integrate the optimization problem in \eqref{eq"qpcvar} for risk estimate as part of the optimal control problem \eqref{eq:cvarcbf}, rather than just listing the constraints therein. Algorithm\ref{alg:DR-CBF} summarizes the steps for the distributionally robust control barrier function.
\begin{algorithm}[H]
\caption{Distributionally Robust Control Barrier Function} \label{alg:DR-CBF}
\begin{algorithmic}[1]
\State \textbf{Require:} $\alpha, \lambda,$ samples 
\While{Not converged}
\State Initialize $x$, $w^{N_s}$, $h(x,w)$
\State Calculate $\mathrm{CVaR}$ using \eqref{eq:1stcase} or \eqref{eq:2ndcase} 
\State Get $\frac{\partial \mathrm{CVaR}_\alpha(h(x,w))}{\partial x}$ by backpropagating through the QP using differentiable convex programming (\Cref{sec:DP})
\State Solve for optimal input $u^*$ using the QP in \eqref{eq:cvarcbf} with $\mathrm{CVaR}$ and its derivative $\frac{\partial \mathrm{CVaR}_\alpha(h(x,w))}{\partial x}$
\State Update $x \leftarrow x + f(x,u) \Delta t $
\If{$x == x_{\text{final}}$}
\State\textbf{Break}
\EndIf
\EndWhile
\end{algorithmic}
\end{algorithm}

\subsection{High-Order System: an Approximate Method} 

Higher-order control barrier functions are an extension of traditional control barrier functions used for higher-order systems. HOCBF incorporates higher-order derivatives of the system's states which allows the consideration of more complex dynamics. Hence, handling more intricate safety requirements and enabling systems to avoid undesirable behaviors. Using the same approach for a higher-order system proves challenging due to the need for two successive differentiation through the Linear programs \eqref{eq:1stcase}and \eqref{eq:2ndcase}. However, we can get a good approximation by differentiating $h(x,w)$ analytically first to get the first layer of the HOCBF \ref{def:HOCBF}, then we can calculate the $\mathrm{CVaR}(\dot{h}(x,w)$ and backpropagate to get the gradients.
We can construct the first Barrier analytically: 
\begin{equation}\label{Hcbfform}
\psi(x,w)= \dot{h}(x,w)+\mathrm{\kappa} (\dot{h}(x,w)) 
\end{equation}
then we can calculate the $\mathrm{CVaR}$ using: 
\begin{equation}\label{eq"HOqpcvar}
\sup _{p \in \mathcal{P}} \operatorname{CVaR}_\alpha^p(\psi(x,w)) = \min _\eta\left\{\eta+\frac{1}{1-\alpha} \frac{1}{N_s} \sum_{m \in\left[N_s\right]} \sup _{w}\left\{[\psi(x,w)-\eta]_{+}-\lambda\left\|w-w^m\right\|\right\}\right\}
\end{equation}
It can be further integrated t in the optimal control problem as in \eqref{eq:cvarcbf} in the following way
\begin{equation}\label{eq:HOcvarcbf}
\min _{u}\left\{J\left(x, u\right) \mid \text { s.t. } \frac{\partial \mathrm{CVaR}_\alpha(\psi(x,w))}{\partial x}\left(f\left(x\right)+g\left(x\right) u\right)+\kappa\left(\operatorname{CVaR}_\alpha(\psi(x,w))\right) \leq 0, \forall m \in\left[N_s\right]\right\}
\end{equation}
Although the resulting CBF is an approximation as $\mathrm{CBC} \circ \mathrm{CVaR}_\alpha \circ \mathrm{CBC} \circ h(x,w)$ instead of $\mathrm{CBC} \circ \mathrm{CBC} \circ \mathrm{CVaR}_\alpha \circ h(x,w)$, its performance was comparable to the first order DR-CBF. We defer the exact method for high-order systems, which requires higher-order differentiable convex programming techniques, to future works.

\section{Simulations and Results}
In this section, we assess the performance of DR-CBF in several scenarios involving first-order systems and second-order systems. Our approach is then compared to a conventional CBF approach while keeping all other configurations identical. The advantages of the proposed DR-CBF on maintaining safety under distributional shift are presented. 
Note that in both these simulations, we use the first approximation of $\mathrm{CVaR}$\eqref{eq:1stcase}, for its computational ease and compromise between optimality and robustness. 
\subsection{Dubins Car: A First-Order System Case Study}
To evaluate the DR-CBF, we used the first-order Dubins car environment with the following kinematics:
\begin{equation}\label{dubinskinematics}
\left(\begin{array}{c}
\dot{x} \\
\dot{y} \\
\dot{\theta}
\end{array}\right)=\left[\begin{array}{ccc}
\cos \theta & -\sin \theta & 0 \\
\sin \theta & \cos \theta & 0 \\
0 & 0 & 1
\end{array}\right]\left(\begin{array}{c}
v_x \\
v_y \\
\omega
\end{array}\right),
\end{equation}
where $v_x$ and $v_y$ are the velocities along the $x$ and $y$ axes of the car's frames, $\theta$ is the heading angle, and $\omega$ is the angular velocity. 

In order to go from an initial state $r_0 = [x_0, y_0, \theta_0]^T$ to a final state $r_f = [x_f, y_f, \theta_f]^T$, we use a Lyapunov function $\frac{1}{2}(r-r_f)^2$, resulting in $J(r,u) = (r-r_f)u + \frac{1}{2}(r-r_f)^2$. 
We describe the safe region by the area outside a circular obstacle in the middle of the car's trajectory and an additive noise $h(r,w) = \rho^2 - \left\|r-r_{obs}\right\|_2^2 + w \leq 0 $
To keep the QP in \eqref{eq:cvarcbf} in a standard Control Lyapunov Function(CLF) form, we rewrite it as an explicit Quadratic Program and integrate the DR-CBF:
 \begin{equation}\label{eq:clfqp}
\begin{aligned}
\min _{u\in[\underline{u}, \bar{u}], \delta} &\ \ \  u^TQu + q^T\delta^2 \\
\text { s.t. }&\ \ \   (r-r_f)u + \frac{1}{2}(r-r_f)^2\leq \delta \\
&\ \ \  \frac{\partial \mathrm{CVaR}_\alpha(h(x,w))}{\partial x}\left(f\left(x\right)+g\left(x\right) u\right)+\kappa\left(\operatorname{CVaR}_\alpha(h(x,w))\right) \leq 0
\end{aligned}
\end{equation}
where $\delta$ is a relaxation factor for the CLF, to allow some divergence from reaching the final point when the safety of the car is compromised and the CBF needs to take over.


\begin{figure}[thpb]
  \centering
  \includegraphics[scale=0.7]{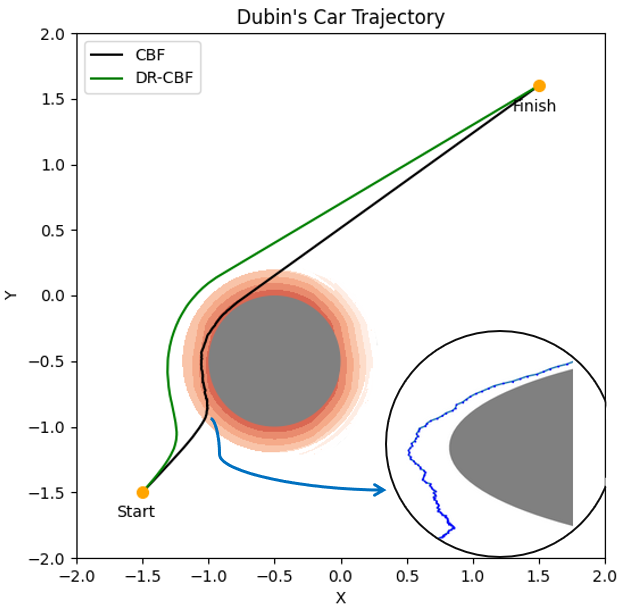}
  \caption{The normal CBF trajectory crosses the noisy region and fluctuates while the DR-CBF avoids the noisy region by a safe margin.}
  \label{fig:Dubinstraj}
\end{figure}
\Cref{fig:Dubinstraj} presents trajectories for the standard CBF and our distributionally robust CBF, while the shaded area shows the contours of the circumference noise. The standard CBF successfully avoids the obstacle but stays amid the noisy region, resulting in a fluctuated trajectory affected by the noise. On the other hand, the DR-CBF takes a more conservative trajectory avoiding the noisy region as well.

\subsection{Quadcopter: A Second-Order System Case Study}
The approximate method for high-order systems was demonstrated on a 2D Quadcopter environment, with the following kinematics: 
\begin{equation}\label{Quadkinematics}
\begin{aligned}
    \ddot{x} &= \frac{T_r + T_l}{m}\sin{\theta} \\
    \ddot{y} &= \frac{T_r + T_l}{m}\cos{\theta} - g\\
    \ddot{\theta} &= (T_r - T_l)\frac{L}{J} 
\end{aligned}
\end{equation}
where $x$ is the horizontal distance of the quadcopter's frame, $y$ is the vertical distance, $\theta$ is the orientation of the quadcopter, $T_r, T_l$ are the right and left rotors thrust/control inputs, $L$ is the arm length of the quadcopter, and $J$ is its moment of inertia.
We simulate a circular trajectory reference problem, where the quadcopter would come across four obstacles along its circular path, with added noise to their circumference $h(r,w) = \rho^2 - \left\|r-r_{obs}\right\|_2^2 + w \leq 0 $, with $r=[x,y]$. This will require twice differentiations to get the control input in the CBF as this is a second-order system. We evaluate the performance of the CBF and DR-CBF in tracking the trajectory and avoiding the obstacles by a safe margin.

\begin{figure}[thpb]
  \centering
  \includegraphics[scale=0.5]{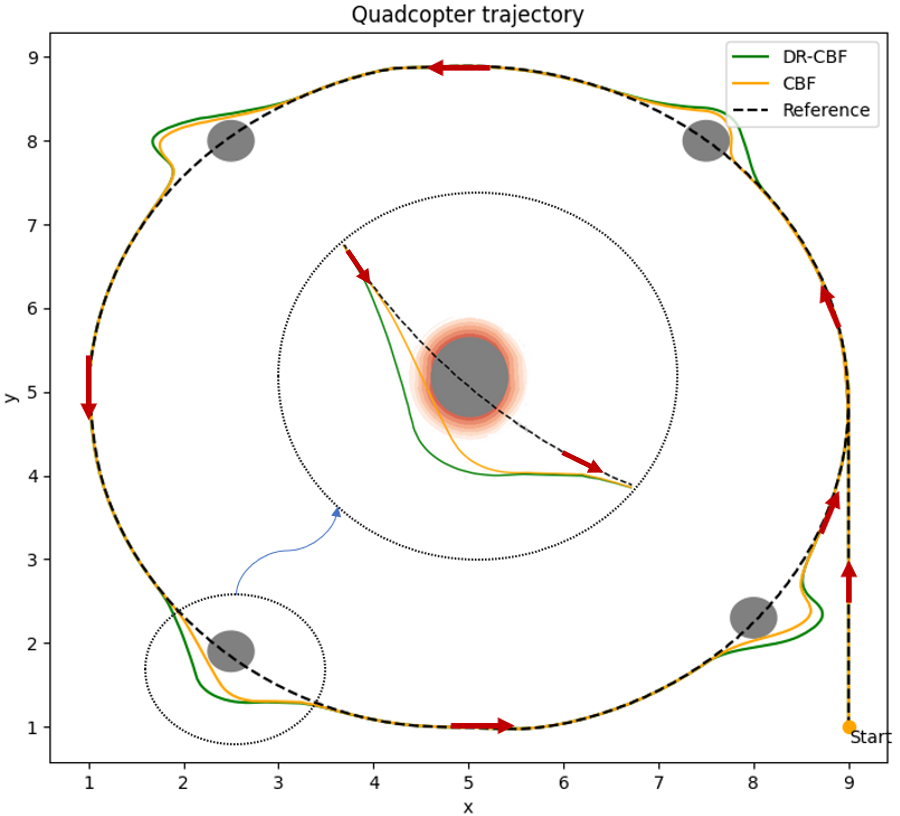}
  \caption{The DR-CBF follows a more conservative trajectory to avoid the potentially unsafe region compared to normal CBF.}
  \label{fig:quadtraj}
\end{figure}
In \Cref{fig:quadtraj}, it is demonstrated that the CBF stays close to the obstacle crossing the noisy region (magnified picture) while the DR-CBF starts steering clear of the obstacle earlier to avoid the noise, and keeps a safer distance from it. Due to the agility of the Quadcopter, we see a throwing motion after avoiding the obstacle for both algorithms, which reflects a slight delay in returning back to the reference trajectory.\Cref{tab:parameters} summarizes the values for the coefficients used in each problem.

\section{Conclusion}

In this paper, we devise a distributionally robust control barrier function for stochastic constraints, using the conditional value at risk, and convex differentiable programming. The proposed framework results in a safer and more robust variant of CBF.
In further work, we want to explore methods for the following settings. (1) More complex, non-additive, and multidimensional noise, which leads to a harder problem for solving the supremum problem. (2) Solving the optimization under distributional constraints exactly with the primal-dual method without taking the dual variable as a constant penalty coefficient. This requires further work to re-cast the optimization over the dual variable in a tractable way. (3) Exact methods for higher-order systems, which require higher-order differentiable convex programming techniques.

\section*{Appendix}
\begin{table}[h]
\caption{Parameters}
\label{tab:parameters}
\begin{center}
\begin{tabular}{l|l|l}
\hline
\hline
Parameters & Description & Value\\
\hline
\hline
$\alpha$  & Confidence level & $0.95$ \\
\hline
$\lambda$ & Penalty coefficient for Wasserstein metric  & $1$ \\
\hline
$\mu$ & Mean of samples & $0.0$ \\
\hline
$\sigma$ & Variance of samples & $0.1$ \\
\hline
\hline
$\mathrm{\kappa_1}$ & Dubins linear kappa coefficient & $1$ \\
\hline
$\mathrm{\kappa_2}$ & Quadcopter linear kappa coefficient & $6$ \\
\hline
$\mathrm{\kappa_3}$ & Quadcopter linear kappa coefficient & $2$ \\
\hline
$\mathrm{\kappa_4}$ & Quadcopter linear kappa coefficient & $12$ \\
\hline
$\mathrm{\kappa_5}$ & Quadcopter linear kappa coefficient & $4$ \\
\hline
$\mathrm{\kappa_6}$ & Quadcopter linear kappa coefficient & $15$ \\
\hline
$\mathrm{\kappa_7}$ & Quadcopter linear kappa coefficient & $5$ \\
\hline
$\mathrm{\kappa_8}$ & Quadcopter linear kappa coefficient & $8$ \\
\hline
$\mathrm{\kappa_9}$ & Quadcopter linear kappa coefficient & $5$ \\
\hline
\end{tabular}
\end{center}
\end{table}

\bibliography{sample}

\begin{thebibliography}{40}
\newcommand{\enquote}[1]{``#1''}
\providecommand{\natexlab}[1]{#1}
\providecommand{\url}[1]{\texttt{#1}}
\providecommand{\urlprefix}{URL }
\expandafter\ifx\csname urlstyle\endcsname\relax
  \providecommand{\doi}[1]{\discretionary{}{}{}https://doi.org/#1}\else
  \providecommand{\doi}[1]{\discretionary{}{}{}\urlstyle{rm}\url{https://doi.org/#1}}\fi

\bibitem[{Mot(2023)}]{Motor}
\enquote{Motor vehicle registered,} , 2023.
\newblock
  \urlprefix\url{https://www.statista.com/statistics/183505/number-of-vehicles-in-the-united-states-since-1990/},
  accessed on May 25, 2023.

\bibitem[{ind(2023)}]{industrial}
\enquote{Industrial-robot-market-size-worldwide,} , 2023.
\newblock
  \urlprefix\url{https://www.statista.com/statistics/728530/industrial-robot-market-size-worldwide/#:~:text=In\%202020\%2C\%20the\%20size\%20of,surpass\%20165\%20billion\%20U.S.\%20dollars./},
  accessed on May 25, 2023.

\bibitem[{USA(2021)}]{USAF}
\enquote{Department of the Air Force Role in Joint All Domain Operations
  (JADO). Air Force Doctrine Publication (AFDP) 3-99. Maxwell Air Force Base,
  AL,} , 2021.
\newblock
  \urlprefix\url{https://www.doctrine.af.mil/Doctrine-Publications/AFDP-3-99-DAF-Role-in-Jt-All-Domain-Ops-JADO/},
  accessed: 2022-10-19.

\bibitem[{Van~Parys et~al.(2015)Van~Parys, Kuhn, Goulart, and
  Morari}]{van2015distributionally}
Van~Parys, B.~P., Kuhn, D., Goulart, P.~J., and Morari, M.,
  \enquote{Distributionally robust control of constrained stochastic systems,}
  \emph{IEEE Transactions on Automatic Control}, Vol.~61, No.~2, 2015, pp.
  430--442.

\bibitem[{Hakobyan and Yang(2021)}]{hakobyan2021wasserstein}
Hakobyan, A., and Yang, I., \enquote{Wasserstein distributionally robust motion
  control for collision avoidance using conditional value-at-risk,} \emph{IEEE
  Transactions on Robotics}, Vol.~38, No.~2, 2021, pp. 939--957.

\bibitem[{Bahari~Kordabad et~al.(2022)Bahari~Kordabad, Wisniewski, and
  Gros}]{bahari2022safe}
Bahari~Kordabad, A., Wisniewski, R., and Gros, S., \enquote{Safe Reinforcement
  Learning Using Wasserstein Distributionally Robust MPC and Chance
  Constraint,} 2022.

\bibitem[{Coulson et~al.(2021)Coulson, Lygeros, and
  D{\"o}rfler}]{coulson2021distributionally}
Coulson, J., Lygeros, J., and D{\"o}rfler, F., \enquote{Distributionally robust
  chance constrained data-enabled predictive control,} \emph{IEEE Transactions
  on Automatic Control}, Vol.~67, No.~7, 2021, pp. 3289--3304.

\bibitem[{Coppens and Patrinos(2021)}]{coppens2021data}
Coppens, P., and Patrinos, P., \enquote{Data-driven distributionally robust MPC
  for constrained stochastic systems,} \emph{IEEE Control Systems Letters},
  Vol.~6, 2021, pp. 1274--1279.

\bibitem[{Yang(2020)}]{yang2020wasserstein}
Yang, I., \enquote{Wasserstein distributionally robust stochastic control: A
  data-driven approach,} \emph{IEEE Transactions on Automatic Control},
  Vol.~66, No.~8, 2020, pp. 3863--3870.

\bibitem[{Ames et~al.(2016)Ames, Xu, Grizzle, and Tabuada}]{ames2016control}
Ames, A.~D., Xu, X., Grizzle, J.~W., and Tabuada, P., \enquote{Control barrier
  function based quadratic programs for safety critical systems,} \emph{IEEE
  Transactions on Automatic Control}, Vol.~62, No.~8, 2016, pp. 3861--3876.

\bibitem[{Long et~al.(2022)Long, Yi, Cortes, and Atanasov}]{long2022safe}
Long, K., Yi, Y., Cortes, J., and Atanasov, N., \enquote{Safe and stable
  control synthesis for uncertain system models via distributionally robust
  optimization,} \emph{arXiv preprint arXiv:2210.01341}, 2022.

\bibitem[{Madry et~al.(2017)Madry, Makelov, Schmidt, Tsipras, and
  Vladu}]{madry2017towards}
Madry, A., Makelov, A., Schmidt, L., Tsipras, D., and Vladu, A.,
  \enquote{Towards deep learning models resistant to adversarial attacks,}
  \emph{arXiv preprint arXiv:1706.06083}, 2017.

\bibitem[{L{\"u}tjens et~al.(2020)L{\"u}tjens, Everett, and
  How}]{lutjens2020certified}
L{\"u}tjens, B., Everett, M., and How, J.~P., \enquote{Certified adversarial
  robustness for deep reinforcement learning,} \emph{Conference on Robot
  Learning}, PMLR, 2020, pp. 1328--1337.

\bibitem[{Liu et~al.(2022)Liu, Guo, Cen, Zhang, Tan, Li, and
  Zhao}]{liu2022robustness}
Liu, Z., Guo, Z., Cen, Z., Zhang, H., Tan, J., Li, B., and Zhao, D.,
  \enquote{On the robustness of safe reinforcement learning under observational
  perturbations,} \emph{arXiv preprint arXiv:2205.14691}, 2022.

\bibitem[{Tessler et~al.(2019)Tessler, Efroni, and Mannor}]{tessler2019action}
Tessler, C., Efroni, Y., and Mannor, S., \enquote{Action robust reinforcement
  learning and applications in continuous control,} \emph{International
  Conference on Machine Learning}, PMLR, 2019, pp. 6215--6224.

\bibitem[{Mankowitz et~al.(2019)Mankowitz, Levine, Jeong, Shi, Kay,
  Abdolmaleki, Springenberg, Mann, Hester, and
  Riedmiller}]{mankowitz2019robust}
Mankowitz, D.~J., Levine, N., Jeong, R., Shi, Y., Kay, J., Abdolmaleki, A.,
  Springenberg, J.~T., Mann, T., Hester, T., and Riedmiller, M.,
  \enquote{Robust reinforcement learning for continuous control with model
  misspecification,} \emph{arXiv preprint arXiv:1906.07516}, 2019.

\bibitem[{Brunke et~al.(2022)Brunke, Greeff, Hall, Yuan, Zhou, Panerati, and
  Schoellig}]{brunke2022safe}
Brunke, L., Greeff, M., Hall, A.~W., Yuan, Z., Zhou, S., Panerati, J., and
  Schoellig, A.~P., \enquote{Safe learning in robotics: From learning-based
  control to safe reinforcement learning,} \emph{Annual Review of Control,
  Robotics, and Autonomous Systems}, Vol.~5, 2022, pp. 411--444.

\bibitem[{Achiam et~al.(2017)Achiam, Held, Tamar, and
  Abbeel}]{achiam2017constrained}
Achiam, J., Held, D., Tamar, A., and Abbeel, P., \enquote{Constrained policy
  optimization,} \emph{Proceedings of the 34th International Conference on
  Machine Learning-Volume 70}, JMLR. org, 2017, pp. 22--31.

\bibitem[{Zhang et~al.(2020)Zhang, Chen, Xiao, Li, Liu, Boning, and
  Hsieh}]{zhang2020robust1}
Zhang, H., Chen, H., Xiao, C., Li, B., Liu, M., Boning, D., and Hsieh, C.-J.,
  \enquote{Robust deep reinforcement learning against adversarial perturbations
  on state observations,} \emph{Advances in Neural Information Processing
  Systems}, Vol.~33, 2020, pp. 21024--21037.

\bibitem[{Sun et~al.(2022)Sun, Kim, and How}]{sun2022romax}
Sun, C., Kim, D.-K., and How, J.~P., \enquote{ROMAX: Certifiably Robust Deep
  Multiagent Reinforcement Learning via Convex Relaxation,} \emph{2022
  International Conference on Robotics and Automation (ICRA)}, IEEE, 2022, pp.
  5503--5510.

\bibitem[{Ren and Majumdar(2022)}]{ren2022distributionally}
Ren, A.~Z., and Majumdar, A., \enquote{Distributionally robust policy learning
  via adversarial environment generation,} \emph{IEEE Robotics and Automation
  Letters}, Vol.~7, No.~2, 2022, pp. 1379--1386.

\bibitem[{Morrison et~al.(2020)Morrison, Corke, and Leitner}]{morrison2020egad}
Morrison, D., Corke, P., and Leitner, J., \enquote{Egad! an evolved grasping
  analysis dataset for diversity and reproducibility in robotic manipulation,}
  \emph{IEEE Robotics and Automation Letters}, Vol.~5, No.~3, 2020, pp.
  4368--4375.

\bibitem[{Wang et~al.(2019)Wang, Tseng, Li, Jiang, Guo, Danielczuk, Mahler,
  Ichnowski, and Goldberg}]{wang2019adversarial}
Wang, D., Tseng, D., Li, P., Jiang, Y., Guo, M., Danielczuk, M., Mahler, J.,
  Ichnowski, J., and Goldberg, K., \enquote{Adversarial grasp objects,}
  \emph{2019 IEEE 15th International Conference on Automation Science and
  Engineering (CASE)}, IEEE, 2019, pp. 241--248.

\bibitem[{Xu et~al.(2022)Xu, Huang, Niu, Kumar, Qiu, Fang, Lee, Qi, Lam, Li
  et~al.}]{xu2022group}
Xu, M., Huang, P., Niu, Y., Kumar, V., Qiu, J., Fang, C., Lee, K.-H., Qi, X.,
  Lam, H., Li, B., et~al., \enquote{Group Distributionally Robust Reinforcement
  Learning with Hierarchical Latent Variables,} \emph{arXiv preprint
  arXiv:2210.12262}, 2022.

\bibitem[{Yang et~al.(2022)Yang, Zheng, Ratliff, Boots, and
  Smith}]{yang2022stackelberg}
Yang, B., Zheng, L., Ratliff, L.~J., Boots, B., and Smith, J.~R.,
  \enquote{Stackelberg MADDPG: Learning Emergent Behaviors via Information
  Asymmetry in Competitive Games,} 2022.

\bibitem[{Xiao and Belta(2021)}]{xiaohigh}
Xiao, W., and Belta, C., \enquote{High-order control barrier functions,}
  \emph{IEEE Transactions on Automatic Control}, Vol.~67, No.~7, 2021, pp.
  3655--3662.

\bibitem[{Amos and Kolter(2017)}]{amos2017optnet}
Amos, B., and Kolter, J.~Z., \enquote{Optnet: Differentiable optimization as a
  layer in neural networks,} \emph{arXiv preprint arXiv:1703.00443}, 2017.

\bibitem[{Agrawal et~al.(2019{\natexlab{a}})Agrawal, Amos, Barratt, Boyd,
  Diamond, and Kolter}]{diffcvxoptlayer}
Agrawal, A., Amos, B., Barratt, S., Boyd, S., Diamond, S., and Kolter, J.~Z.,
  \enquote{Differentiable convex optimization layers,} \emph{Advances in neural
  information processing systems}, Vol.~32, 2019{\natexlab{a}}.

\bibitem[{Agrawal et~al.(2019{\natexlab{b}})Agrawal, Barratt, Boyd, Busseti,
  and Moursi}]{diffconeprog}
Agrawal, A., Barratt, S., Boyd, S., Busseti, E., and Moursi, W.~M.,
  \enquote{Differentiating through a cone program,} \emph{arXiv preprint
  arXiv:1904.09043}, 2019{\natexlab{b}}.

\bibitem[{Rockafellar et~al.(2000)Rockafellar, Uryasev
  et~al.}]{rockafellar2000optimization}
Rockafellar, R.~T., Uryasev, S., et~al., \enquote{Optimization of conditional
  value-at-risk,} \emph{Journal of risk}, Vol.~2, 2000, pp. 21--42.

\bibitem[{Kleywegt et~al.(2002)Kleywegt, Shapiro, and Homem-de
  Mello}]{kleywegt2002sample}
Kleywegt, A.~J., Shapiro, A., and Homem-de Mello, T., \enquote{The sample
  average approximation method for stochastic discrete optimization,}
  \emph{SIAM Journal on optimization}, Vol.~12, No.~2, 2002, pp. 479--502.

\bibitem[{Sinha et~al.(2017)Sinha, Namkoong, and Duchi}]{sinha2017certifiable}
Sinha, A., Namkoong, H., and Duchi, J., \enquote{Certifiable distributional
  robustness with principled adversarial training,} \emph{arXiv preprint
  arXiv:1710.10571}, Vol.~2, 2017.

\bibitem[{Blanchet and Murthy(2019)}]{blanchet2019quantifying}
Blanchet, J., and Murthy, K., \enquote{Quantifying distributional model risk
  via optimal transport,} \emph{Mathematics of Operations Research}, Vol.~44,
  No.~2, 2019, pp. 565--600.

\bibitem[{Rahimian and Mehrotra(2019)}]{rahimian2019distributionally}
Rahimian, H., and Mehrotra, S., \enquote{Distributionally robust optimization:
  A review,} \emph{arXiv preprint arXiv:1908.05659}, 2019.

\bibitem[{Hu and Hong(2013)}]{hu2013kullback}
Hu, Z., and Hong, L.~J., \enquote{Kullback-Leibler divergence constrained
  distributionally robust optimization,} \emph{Available at Optimization
  Online}, Vol.~1, No.~2, 2013, p.~9.

\bibitem[{Givens and Shortt(1984)}]{givens1984class}
Givens, C.~R., and Shortt, R.~M., \enquote{A class of Wasserstein metrics for
  probability distributions.} \emph{Michigan Mathematical Journal}, Vol.~31,
  No.~2, 1984, pp. 231--240.

\bibitem[{Mohajerin~Esfahani and Kuhn(2018)}]{mohajerin2018data}
Mohajerin~Esfahani, P., and Kuhn, D., \enquote{Data-driven distributionally
  robust optimization using the Wasserstein metric: Performance guarantees and
  tractable reformulations,} \emph{Mathematical Programming}, Vol. 171, No.
  1-2, 2018, pp. 115--166.

\bibitem[{Amos et~al.(2018)Amos, Jimenez, Sacks, Boots, and
  Kolter}]{amos2018differentiable}
Amos, B., Jimenez, I., Sacks, J., Boots, B., and Kolter, J.~Z.,
  \enquote{Differentiable mpc for end-to-end planning and control,}
  \emph{Advances in neural information processing systems}, Vol.~31, 2018.

\bibitem[{Agrawal et~al.(2019{\natexlab{c}})Agrawal, Barratt, Boyd, Busseti,
  and Moursi}]{agrawal2019differentiating}
Agrawal, A., Barratt, S., Boyd, S., Busseti, E., and Moursi, W.~M.,
  \enquote{Differentiating through a cone program,} \emph{arXiv preprint
  arXiv:1904.09043}, 2019{\natexlab{c}}.

\bibitem[{Agrawal et~al.(2020)Agrawal, Barratt, Boyd, and
  Stellato}]{agrawal2020learning}
Agrawal, A., Barratt, S., Boyd, S., and Stellato, B., \enquote{Learning convex
  optimization control policies,} \emph{Learning for Dynamics and Control},
  PMLR, 2020, pp. 361--373.

\end{thebibliography}

\end{document}